%% file: main.tex
\documentclass[letterpaper,10pt,conference]{IEEEtran}
\usepackage[letterpaper,left=0.765in,right=0.765in,top=0.75in,bottom=0.85in,columnsep=0.2in]{geometry}
\renewcommand{\IEEEtitletopspaceextra}{0.18in}
\IEEEoverridecommandlockouts
\usepackage{cite}
\usepackage{amsmath,amssymb,amsfonts}
\usepackage{graphicx}
\usepackage{booktabs,multirow,textcomp,array}
\usepackage{xcolor}

\title{\fontsize{16}{20}\selectfont\bfseries Same World, Different Knowledge:\\
When Isolated Audits Misjudge World-Model Repairs}
\author{
\IEEEauthorblockN{
Rui Min\IEEEauthorrefmark{1},
Xianyao Li\IEEEauthorrefmark{2},
Fang Xu\IEEEauthorrefmark{2},
Sofiane Lachab\IEEEauthorrefmark{3}, and
Jing Du\IEEEauthorrefmark{2}}
\IEEEauthorblockA{\IEEEauthorrefmark{1}Department of Mechanical and Aerospace Engineering, University of Florida, Gainesville, FL, USA}
\IEEEauthorblockA{\IEEEauthorrefmark{2}Department of Civil and Coastal Engineering, University of Florida, Gainesville, FL, USA}
\IEEEauthorblockA{\IEEEauthorrefmark{3}Molinaroli College of Engineering and Computing, University of South Carolina, Columbia, SC, USA}
\IEEEauthorblockA{Corresponding author: Jing Du (eric.du@essie.ufl.edu)}
}

\begin{document}
\maketitle

\begin{abstract}
A repair favored under an isolated input fault can be inferior when deployed modules share the faulty information. We introduce an information-interface audit for world models, distinguishing fidelity gaps, where exact inputs become estimates, from availability gaps, where inputs are missing. Fixed-weight interventions measure prediction error, input dependence, and paired closed-loop benefit, including dependencies introduced by reconstruction. In simulated quadrotor model predictive control (MPC), coupled, opposite-sign 10\% mass/thrust calibration errors reduce a physics-anchored model's success from 69\% to 8\%; uncertainty training restores 65\%. Wind reconstruction recovers control benefit but inherits calibration dependence. For a positive calibration offset, reconstruction-only corruption favors uncertainty-trained reconstruction, whereas shared corruption favors the baseline. Acceleration diagnostics reveal compensation between reconstruction bias and nominal-model error, also observed with a disturbance observer. Repair selection therefore depends on the information paths used in deployment.
\end{abstract}

\section{Introduction}
A learned world model acts through an information interface. In
simulation, this interface can supply exact state, parameters, and
disturbances; a robot must measure them, estimate them, or act without
them. Internal-model approaches connect prediction to
action~\cite{craik1943nature, sutton1990dyna, ha2018world}. The same information can affect a
prediction through several paths. A calibration value may enter the
dynamics predictor directly and also affect the wind estimate supplied
by an upstream reconstruction module.

This raises a concrete question: \emph{does a repair that improves
control under an isolated information fault still help when that
fault is shared by the deployed modules?} A wind reconstruction module
and its predictor may share calibration inputs. We compare corruption at
one module with corruption at the shared source, testing both in closed loop.

We study two gaps: a \emph{fidelity gap}, where an exact training input
becomes an estimate, and an \emph{availability gap}, where a training
channel cannot be supplied. Mass and thrust calibration inputs illustrate the
former; wind without a wind sensor illustrates the latter.
We intervene on model inputs while fixing the weights, plant, and planner,
then measure prediction error, input dependence, and paired control
benefit. Our quadrotor experiments motivate this distinction: prediction
errors span a 12.6$\times$ range, yet paired tests detect no
success-rate difference under correct inputs.

Our contributions are an information-interface specification, a paired
intervention protocol, and a rule for constructing and comparing
repairs. The specification records where each input comes from, which
modules use it, and how it is supplied at deployment. This makes shared
information an explicit audit target. The protocol distinguishes a model's dependence
on a channel from the control benefit that channel provides.
\emph{Deployment-consistent conditioning} (DCC) supplies candidate
repairs through uncertainty training, withholding, and reconstruction.
The audit measures each repair's benefit under faulty inputs and its
nominal performance cost: any loss under correct inputs.
In our quadrotor experiments, re-auditing the connected modules reverses repair
preference under a positive calibration offset: DCC-trained reconstruction is
favored under reconstruction-only corruption, and the baseline under shared corruption.
A residual disturbance observer (DOB) tests the same interface change
with a different reconstruction mechanism. Fixed-history diagnostics show that
uncertainty training attenuates compensation between reconstruction bias
and calibration error in a nominal acceleration model.

\begin{figure*}[t]
\centering
\includegraphics[width=\textwidth]{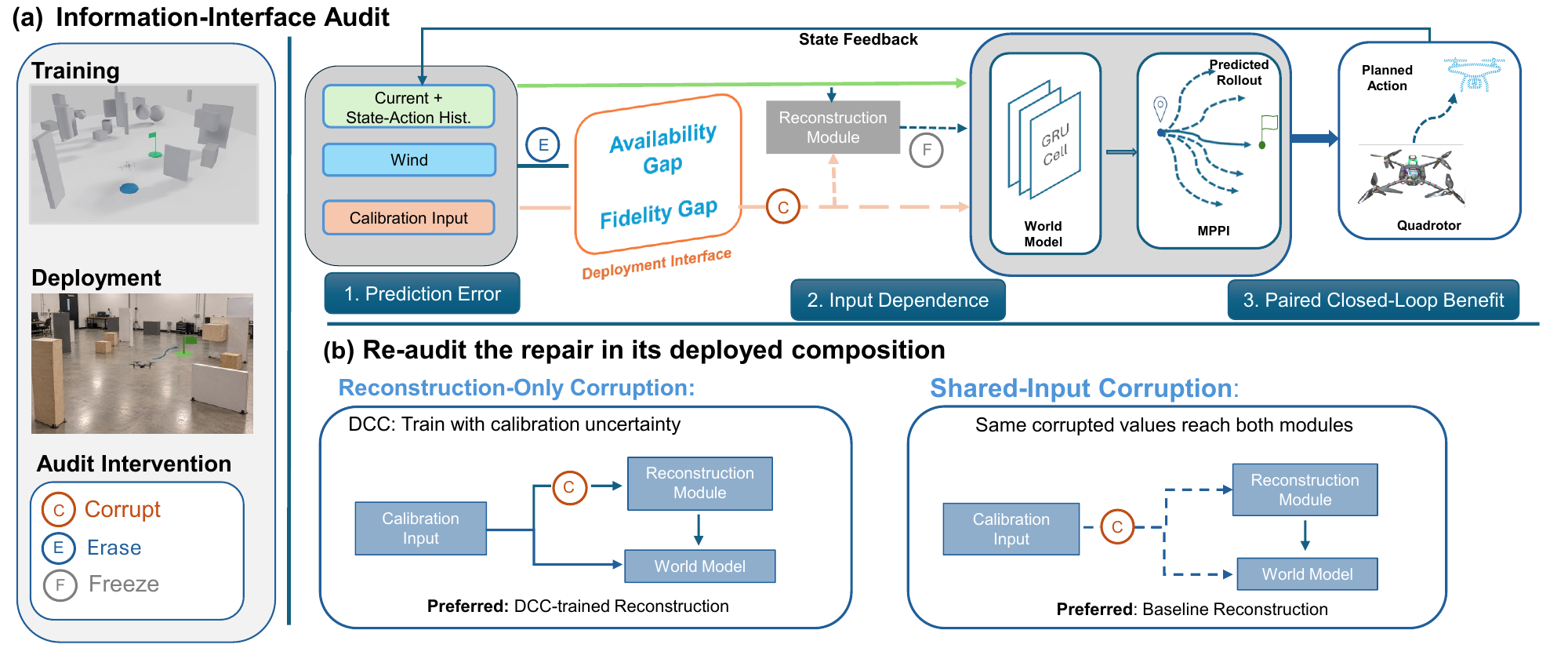}
\caption{The information-interface audit. (a) Fidelity and availability
gaps arise when exact calibration inputs become estimates and wind is not
directly measured. A reconstruction module recovers wind from state--action
history and also uses calibration inputs. C, E, and F mark corruption,
erasure, and freezing; the audit measures prediction error, input dependence,
and paired closed-loop benefit. (b) Re-auditing wind reconstruction in its
deployed composition (Table~\ref{tab:shared}) changes the preference:
at $e=+0.1$, reconstruction-only corruption favors DCC-trained reconstruction, while
shared corrupted calibration favors baseline reconstruction. Scene images:
simulation render (top) and generated deployment illustration (bottom).}
\label{fig:interface}
\end{figure*}

\section{Related Work}
\subsubsection{Objective Mismatch}
Predictive accuracy need not track control quality. Lambert et
al.~\cite{lambert2020mismatch} study this objective mismatch;
Wei et al.~\cite{wei2024unified} survey decision-aware responses.
Smolyanskiy~\cite{smolyanskiy2026checkpoint} selects checkpoints using
structural diagnostics when prediction and control scores diverge.
Model-based policy learning also links model error to downstream
performance~\cite{janner2019mbpo}. We fix predictor weights and change
the information supplied about the same physical system.

\subsubsection{What a World Model Must Reproduce}
Control-oriented world models include locally linear latent
dynamics~\cite{watter2015e2c}, value-prediction networks~\cite{oh2017vpn},
models based on value equivalence~\cite{grimm2020value}, and latent-space agents~\cite{hafner2020dreamer, hansen2024tdmpc2}.
SkyJEPA~\cite{rao2026skyjepa} combines latent prediction, physical decoding,
and sampling-based MPC. Its input-noise tests hold reference trajectories
fixed, and its prober comparisons evaluate alternative decoding modules.

\subsubsection{Physical and Information Variation}
Domain and dynamics randomization vary the simulated plant~\cite{tobin2017dr, peng2018dynrand}.
Adaptation and teacher--student schemes recover unobserved
context~\cite{kumar2021rma, lee2020quadruped, chen2020lbc}.
RMA compares its base policy with and without the adaptation module to
assess that module's contribution~\cite{kumar2021rma}.
Our datasets also use dynamics randomization, but the calibration audit
changes supplied parameters while keeping the plant fixed. DCC corrupts
training inputs while retaining samples and targets, an errors-in-variables
condition~\cite{fuller1987eiv}.

\subsubsection{Attribution and Trust}
DA-LeWM~\cite{wang2026dma} separates latent-state decodability from
candidate-cost ranking. Its ablations compare auxiliary objectives at the
same training budget, using ranking and closed-loop success. These practices motivate
a further comparison: whether a repair preference survives when an
isolated input fault is shared across deployed modules.
Uncertainty-aware control~\cite{chua2018pets}, quadrotor dynamics
modeling~\cite{bauersfeld2021neurobem}, and high-speed flight~\cite{kaufmann2023racing}
improve prediction or control. Our interventions examine dependence on
the information available within a fixed controller.

\subsubsection{Disturbance Compensation}
Joint disturbance-model and observer
design already addresses unmeasured disturbances and plant/model
mismatch in offset-free MPC~\cite{pannocchia2007combined}.
Fixed-time observers have been evaluated on real
quadrotors~\cite{xu2024fxtdo}, and adaptive MPC combines prediction
with online compensation under changing
conditions~\cite{pereida2018adaptive}. Our GRU, GRU+DCC, and DOB comparison
tests how isolated and shared input faults change reconstruction preference.

\section{A Framework for Auditing Information Interfaces}
\label{sec:framework}
A model's information interface describes how information reaches it.
Figure~\ref{fig:interface} connects input dependence, control benefit,
and the dependencies introduced by a repair.
\par

\subsection{Classify the Channel at Deployment}
We describe each module by the information delivered to it. Let
$c=(c_1,\ldots,c_K)$ collect the inputs from its conditioning channels,
and let $\theta$ denote the underlying physical context. The input
distributions $q_\mathrm{train}(c\mid\theta)$ and
$q_\mathrm{deploy}(c\mid\theta)$ may differ even when $\theta$ is fixed.
A predictor with weights $\phi$ uses these inputs and a candidate action
sequence $U$ to produce a rollout $\hat X=F_\phi(c,U)$.

We study two gaps relative to an accurate-input reference.
A \emph{fidelity gap} replaces truth with an estimate; an
\emph{availability gap} removes a channel the trained module expects.
For each channel, we record its source, the components that use it, and
its deployment conditions, including latency, update rate, and error
correlation. At decision time $t$, reconstruction uses only the measurements,
executed actions, and estimates already in the controller's information set
$\mathcal F_t$, excluding future states and disturbances.
For a shared source, we trace every use of its output, including input
normalization, analytic model branches, and reconstruction modules.

\subsection{Separate Dependence from Control Benefit}
Diagnosis fixes weights, plant, task, planner, initial conditions,
and per-scene exogenous disturbance processes. \emph{Corruption}
changes an input's value or timing while its physical source stays
unchanged. \emph{Erasure} replaces a channel by a specified default.
\emph{Freezing} holds its value after a specified time to test the
benefit of continued updates within either gap. Each intervention
specifies its onset, persistence, and affected consumers.

Closed-loop trials pair initial conditions and disturbances; subsequent
histories can diverge with the chosen actions. We average within-scene
success differences and compare completion time, terminal goal distance,
and minimum clearance. Offline probes hold histories and candidate actions fixed.

Prediction error compares rollouts with ground truth on common windows.
An input-perturbation probe instead measures the response to an input
change with all others fixed. Replacing $s$ by $s'$ gives
$A_s(H)=\mathrm{median}\,\|\hat p_{t+H}(s')-\hat p_{t+H}(s)\|$.
The median is taken over trajectory windows.
Paired closed-loop outcomes measure the intervention's control consequence.
Together, the probe and closed-loop comparison distinguish a channel
that is barely used from one that changes predictions without a detected
net benefit. Probe comparisons must account for perturbation scale.

\subsection{Propose Repairs, Then Audit Their Composition}
For estimated inputs, DCC perturbs training inputs under a specified
uncertainty model while preserving physical samples and targets.
For unavailable inputs, withhold the channel during training or
reconstruct it from obtainable channels. Erasure diagnoses an existing
dependency; withholding retrains without it. Matched repair comparisons
preserve architecture and loss and evaluate both faulty and correct inputs.

With training data $\mathcal D$ and rollout loss $\ell$,
uncertain-input training minimizes
\begin{equation}
\begin{split}
\mathcal L_{\mathrm{DCC}}(\phi)
&=\mathbb E_{(c,U,X)\sim\mathcal D}\\
&\quad\mathbb E_{\epsilon\sim q_\epsilon}
\big[\ell(F_\phi(T_\epsilon(c),U),X)\big].
\end{split}
\label{eq:dcc}
\end{equation}
Here $T_\epsilon$ changes the conditioning inputs, not the target $X$;
$q_\epsilon$ specifies noise magnitude, persistence, and cross-channel
dependence. Withholding uses the same replacement in training and
evaluation; reconstruction supplies the unchanged downstream controller.

The rule is recursive: a reconstruction module
$\hat c_s=\mathcal R(c_{\mathcal O})$ depends on its own available inputs
$c_{\mathcal O}$, which require the same audit as its output.
Compare each repair with its baseline under both isolated and shared
faults. The decision record (Table~\ref{tab:record}) pairs fault benefits
with nominal costs and identifies the fault frequency needed to offset them.

For a new system, rank uncertain sources first by consumer count, then
by probe response to perturbations scaled to expected deployment uncertainty.
Start with corruption at the repaired module and
at its shared source; add other consumer-specific faults for attribution.

\begin{figure*}[t]
\centering
\includegraphics[width=\textwidth]{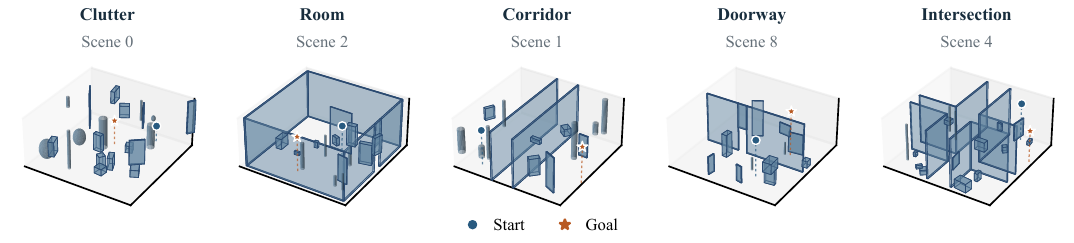}
\caption{Representative held-out scenes: the first library entry of each
evaluation topology. Scene IDs follow the frozen library. Blue circles
and orange stars mark start and goal; transparency reveals the geometry.}
\label{fig:scenes}
\end{figure*}

\section{Experimental Setup: Quadrotor MPC}
\label{sec:setup}
The experiments use a common navigation system for prediction,
calibration, and wind comparisons. Its conditioning paths are shown
in Fig.~\ref{fig:interface}.

\subsection{Control Architecture and Information Interface}
We compare four predictors: a \emph{differentiable kinematic integrator}
(\emph{DKI}), an analytic double-integrator with rate-loop lag; an unanchored
learned predictor (\emph{WM-U}); a physics-anchored model that adds learned
residual accelerations to DKI (\emph{WM-A});
and an analytic model supplied with true physical parameters
(\emph{true-parameter model}).

The state comprises position, velocity, a rotation matrix, and angular
velocity; the action is collective thrust and body-rate command.
Control runs at 20\,Hz, using 12 steps of state/action history to predict
20 steps ahead. At each replanning step, a gated recurrent unit (GRU)
encodes the executed history once. All candidate rollouts start from the
same current state and history encoding. At each rollout step, the
candidate action is embedded and passed to a GRU cell to update the
latent state. A multilayer perceptron (MLP) with two hidden layers uses
this latent state, the predicted physical state, and the action to
compute the next state.

WM-A outputs bounded residual linear and angular
accelerations, which enter the DKI update. WM-U directly predicts
position, velocity, and angular-velocity increments, together with a
rotation increment composed with the current attitude. It preserves
rotation geometry while removing the DKI dynamics branch.

Physical mass and thrust scale are $m,T$; the supplied calibration
inputs are $\tilde m,\tilde T$.
The tilde identifies the supplied value, whether correct, estimated, or
deliberately perturbed; the physical parameters remain fixed during an
input intervention.
Both predictors use 19-dimensional base features (velocity,
rotation matrix, angular velocity, hover-normalized thrust, and
commanded body rates), omitting absolute position.
The calibration input $\tilde m$ normalizes thrust in the history and rollout.
WM-A also supplies $\tilde m,\tilde T$ to DKI; WM-U appends
both to the base vector, giving 21 features. Calibration interventions
cover these paths together.

Both use a 32-dimensional latent state and MLPs with two hidden
layers: 64 units per layer in WM-A and 128 in WM-U. The complete
models contain 17k and 35k parameters, respectively. Training initializes
the recurrent state from history and minimizes normalized state error
over the rollout. Each prediction becomes the input to the next step;
future ground-truth states are used only as targets. We use AdamW
(learning rate $10^{-3}$, batch 256) for 10 epochs (WM-A) or 30 epochs
(WM-U). The two architectures therefore differ in capacity and training
duration. Within each model, DCC preserves
architecture, data, loss, and training budget.
WM-A, WM-U, their +DCC variants, and the clearance model are trained on
our own simulation data: 7,150 episodes of up to 6\,s collected in
Isaac~Sim with a waypoint controller and injected action perturbations
(25,684 32-step windows before an episode-wise training/validation split),
under per-episode randomization of mass, thrust
and torque scale ($\pm50\%$), motor time constant, inertia, and drag
coefficients. The wind study adds a 14,891-episode pool with randomized
wind: predictors with and without a wind input train on an equal mix of
both pools, and the reference reconstruction module and its DCC variant
on the wind pool alone.

The true-parameter model uses ground-truth parameters in replicated
inner-loop and aerodynamic equations. Integration is separate from PhysX,
so its transitions can differ from the simulator's. All predictors are
evaluated on the same trajectory windows.

A separate model estimates obstacle clearance from depth observations
for the planner. Model
predictive path integral control (MPPI) scores 512 candidates over a one-second horizon using goal,
clearance-risk, and speed costs. Its planning core runs at 42\,Hz on
the target flight computer, excluding perception, state estimation, and wind reconstruction.

Free-space trials use ground-truth clearance for environment boundaries;
obstacle trials use the depth model unless a ground-truth-clearance control
is explicitly marked. Free-space trials reduce obstacle-planning
difficulty, while obstacle trials test the same input interventions
during navigation.

State and history come from the simulator; a supplementary evaluation
adds Gaussian noise and delay to these observations. Wind enters through
linear aerodynamics. When a wind input is supplied, its contribution to
predicted acceleration is
$R(c_{\mathrm{lin}}\circ R^\top w)/\tilde m$, using fixed nominal
drag coefficients and the supplied mass $\tilde m$. Current ground-truth
wind and reconstructed wind are each held constant over the planning horizon;
future ground-truth wind is not supplied in closed loop.

\subsection{Interventions and Evidence}
Training, selection, and held-out evaluation use disjoint geometry libraries.
Unless stated otherwise, success rates use the held-out library of
100 scenes spanning
clutter, rooms, corridors, doorways, and intersections, with or without
obstacles. Figure~\ref{fig:scenes} shows one example per topology.
Success requires reaching within 0.5\,m of the goal in
12\,s; contact ends a run as a crash, and the remainder are timeouts.
Scenes and disturbance seeds are paired across interventions.
The reference reconstruction module (\emph{GRU}) and GRU+DCC in
Table~\ref{tab:shared} train on a separate randomized collection pool.
Its wind streams derive from collection seed and episode index, whereas
evaluation uses wind level and scene index. Neither module trains on
the scene-indexed evaluation trajectories. The trajectory-trained variant
in Table~\ref{tab:wind} instead uses selection scenes 0--69; its wind
streams recur at those indices in the held-out geometry library.
Table~\ref{tab:wind} therefore also reports its unseen-wind subset, indices
70--99. Full stream mappings are in the reproduction record.

Calibration corruption sets $\tilde m=m(1+e)$ and
$\tilde T=T(1-e)$. This coupled stress test rescales the implied
thrust acceleration by $(1-e)/(1+e)$. Positive and negative 10\% offsets
therefore imply acceleration changes of roughly $-18\%$ and $+22\%$,
respectively. For WM-A+DCC and WM-U+DCC, training independently perturbs
the two input parameters uniformly within $\pm20\%$. Each perturbation
is sampled once per training window and held fixed throughout it;
the underlying sample and prediction target stay unchanged.
This uniform noise is a robustness assumption, not a measured
deployment-error distribution. The wind study uses sustained,
turbulent, and gust disturbances at four levels: L0 (none), L1 (light),
L2 (moderate), and L3 (strong). The reported comparisons use L0, L2,
and L3.

Scenes are the unit of pairing and inference; tables specify the seed
hierarchy. Single-seed binary comparisons use exact McNemar tests.
For multi-seed comparisons, we average success differences within each
scene, then apply a sign test and bootstrap scenes for percentile
intervals. These intervals describe scene variation conditional on the
trained models and recorded executions. Tables~\ref{tab:calib}
and~\ref{tab:shared} report discordant counts per seed.
Undetected differences are not interpreted as equivalence.

Completion times are compared on jointly successful scenes; terminal
goal distance and minimum clearance use all paired scenes. Full paired
statistics, detection-sensitivity calculations, and training and controller
settings are in the reproduction record.

Code, frozen scene libraries, closed-loop artifacts, and statistical
audit scripts will be released with the paper.

\section{Prediction Accuracy and Closed-Loop Performance}
We first compare predictors with different dynamics representations
under correct inputs. On a common trajectory distribution, their
position errors span a wide range, while paired tests detect no
success-rate difference (Table~\ref{tab:main}). Timing and failure
types vary across the same predictors.

\input{table_main}

All models in this comparison receive correct calibration. The next
comparison keeps each model fixed and changes the values it receives.

\section{The Fidelity Gap: Uncertain Calibration}
\label{sec:calib}
Mass and thrust calibration enter both action normalization and the
rollout model. At deployment these values are estimates; the nominal
training interface supplies them exactly. We perturb the delivered
values while keeping the vehicle and trained weights fixed.

With corrupted calibration, WM-A loses most of its nominal success
under both error signs (Table~\ref{tab:calib}). The unanchored WM-U
also loses substantial success at $e=+0.1$, while its negative-offset
performance stays near nominal. For WM-A, median prediction error
also worsens, but orders the two conditions oppositely to closed-loop success: the lower-error direction
has the lower success rate (Table~\ref{tab:calib}, bottom).
Upper-tail error orders them consistently with success. The two error summaries therefore
lead to different orderings of the same fixed model's input conditions.

\begin{figure}[t]
\centering
\includegraphics[width=\columnwidth]{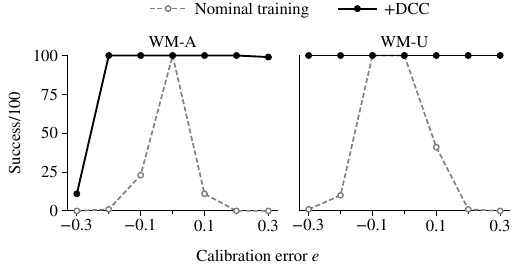}
\caption{Success/100 versus calibration error $e$ on the selection
library in free space (seed 0). WM-U+DCC completes all scenes across
the tested range; WM-A+DCC broadens the nominal peak but still fails at
the most negative offset.}
\label{fig:calib}
\end{figure}

\begin{table}[t]
\caption{Calibration audit on held-out obstacles. Matched comparisons
report rounded three-model-seed mean success/100; differences use unrounded means.
Paired confidence intervals (CIs) bootstrap scene-wise mean differences,
conditional on the trained models. The final panel uses WM-A seed 0:
errors are median 20-step position errors on common held-out trajectories
collected with the true-parameter model; success uses the same model seed.}
\label{tab:calib}
\centering\small
\setlength{\tabcolsep}{2.5pt}
\begin{tabular}{lccc}
\toprule
Model & $e{=}{-}0.1$ & $e{=}0$ & $e{=}{+}0.1$ \\
\midrule
\multicolumn{4}{l}{\emph{Matched WM-A comparison:}}\\
WM-A & 18 & 69 & 8 \\
WM-A+DCC & 68 & 68 & 65 \\
\addlinespace
\multicolumn{4}{l}{\emph{+DCC minus baseline (percentage points):}}\\
Difference & $49.7$ & $-1.0$ & $57.3$ \\
95\% paired CI & $[40.0,59.3]$ & $[-3.3,1.0]$ & $[48.3,66.3]$ \\
Wins:losses & $51{:}0/51{:}1/49{:}1$ & $2{:}2/0{:}2/1{:}2$ & $59{:}0/58{:}0/55{:}0$ \\
\midrule
\multicolumn{4}{l}{\emph{Matched WM-U comparison:}}\\
WM-U & 66 & 70 & 29 \\
WM-U+DCC & 68 & 68 & 69 \\
\addlinespace
\multicolumn{4}{l}{\emph{+DCC minus baseline (percentage points):}}\\
Difference & $2.3$ & $-1.3$ & $40.0$ \\
95\% paired CI & $[-0.7,6.0]$ & $[-3.7,0.7]$ & $[31.7,48.7]$ \\
Wins:losses & $3{:}1/4{:}0/4{:}3$ & $1{:}2/0{:}3/2{:}2$ & $45{:}0/38{:}0/37{:}0$ \\
\midrule
\multicolumn{4}{l}{\emph{WM-A seed 0: prediction and control}}\\
Error (cm) & 74.7 & 8.2 & 61.0 \\
Success/100 & 17 & 69 & 8 \\
\bottomrule
\end{tabular}
\par\smallskip\raggedright
Wins:losses = +DCC-only:baseline-only successes, listed as seeds 0/1/2.
\end{table}

Physical dynamics are unchanged between corrupted-input and nominal trials.
DCC changes the training input channel, exposing the predictor to
uncertain calibration while retaining the physical samples, architecture,
and loss. Most of WM-A's nominal obstacle success is recovered
(Table~\ref{tab:calib}). The wider free-space selection sweep shows
different robustness ranges: WM-U+DCC completes every scene
throughout the tested $\pm0.3$ range, whereas WM-A+DCC still fails at
$e=-0.3$ (Fig.~\ref{fig:calib}).

On common obstacle histories from the selection library, input-perturbation
probes with model seed 0 show how WM-A's dependence changes.
The change in predicted position caused by coupled $e=+0.1$ calibration
corruption falls from about 62 to 9\,cm after repair; the change caused
by history shuffling rises from 2 to 15\,cm. The shuffle is a deliberate dependence probe, not a
sample from a deployment-noise model. These responses indicate a
shift from calibration dependence toward dependence on available history.
Section~\ref{sec:mech} separately tests whether updating that history
improves closed-loop performance.

Repair gains differ with error sign and between architectures;
WM-U is already near nominal success under negative corruption.

With correct calibration and exact state feedback, paired tests detect
no success loss for WM-A+DCC relative to WM-A. Under synthetic state
noise and delay (model seed 0), DCC still improves success under
calibration error. With correct calibration, however, WM-A+DCC succeeds
in 66 scenes versus 73 for WM-A, a detected nominal cost
(0:7 discordant scenes).

\section{The Availability Gap: A Missing Wind Channel}
\label{sec:avail}
Calibration remains available even when inaccurate. Wind has no
corresponding sensor on the vehicle, although simulation can supply
its true value. We compare a wind-conditioned predictor---privileged in
the sense of Chen et al.~\cite{chen2020lbc}---with an otherwise matched
wind-withheld model, using the same physical episodes, split, training
budget, and loss. Both models train on transitions affected by wind;
only the wind-conditioned model receives wind as an input.
Supplying a zero wind input at deployment removes that information without turning off
the physical disturbance.

The wind-conditioned model has lower prediction error than the
wind-withheld model when given ground-truth wind. With a zero wind input
at deployment, it has no detected success advantage over the
wind-withheld model (Table~\ref{tab:wind}), and its prediction error on
common trajectories is higher.

\begin{table}[t]
\caption{Held-out success/100 under wind, with and without obstacles.
Predictor training and deployment wind inputs are listed separately.
Entries without a footnote use seed 0. GRU denotes the collection-trained
reference reconstruction module; the trajectory-trained variant is labeled
separately. At L0, ground-truth wind is zero; those cells reuse the zero-input runs.}
\label{tab:wind}
\centering\small
\setlength{\tabcolsep}{3pt}
\begin{tabular}{@{}llccc@{}}
\toprule
Training & Deployment input & L0 & L2 & L3 \\
\midrule
\multicolumn{5}{@{}l}{\emph{Held-out, free space:}}\\
wind-withheld & none & 100 & 98 & 82.7\textsuperscript{a} \\
wind-conditioned & zero & 100 & 96 & 78.7\textsuperscript{a} \\
wind-conditioned & $\hat w_t$ (trajectory GRU) & 100 & 100 & 96.7\textsuperscript{b} \\
wind-conditioned & $\hat w_t$ (GRU) & 100 & 100 & 100 \\
wind-conditioned & $w_t$ (ground truth) & 100 & 100 & 100 \\
\midrule
\multicolumn{5}{@{}l}{\emph{Held-out, obstacles:}}\\
wind-withheld & none & 69 & 62 & 44 \\
wind-conditioned & zero & 69 & 60 & 39 \\
wind-conditioned & $\hat w_t$ (trajectory GRU) & 59 & 59 & 48.3\textsuperscript{b} \\
wind-conditioned & $\hat w_t$ (GRU) & 70 & 68 & 67.0\textsuperscript{b} \\
wind-conditioned & $w_t$ (ground truth) & 69 & 68 & 65 \\
\bottomrule
\end{tabular}
\par\smallskip\raggedright
\textsuperscript{a}Mean over three world-model seeds;
\textsuperscript{b}mean over three reconstruction seeds, predictor seed 0.
\par\smallskip\centering\footnotesize
\begin{tabular}{@{}lcc@{}}
\toprule
\multicolumn{3}{@{}l}{\emph{L3 unseen-wind subset: success/30, indices 70--99}}\\
Deployment input & Free space & Obstacles \\
\midrule
$\hat w_t$ (trajectory GRU), seeds 0/1/2 & 30/29/30 & 14/15/14 \\
zero, predictor seed 0 & 22 & 8 \\
$w_t$ (ground truth), predictor seed 0 & 30 & 19 \\
\bottomrule
\end{tabular}
\end{table}

% Queue the paired audit tables early enough to keep them with the results.
\input{table_shared_calibration}
\input{table_audit_record}

Withholding allows the predictor to operate without a wind input.
Supplying current ground-truth wind to the same fixed-weight,
wind-conditioned predictor increases success relative to a zero wind input.
This motivates reconstruction from the vehicle's available observations.
The reference module uses a separate GRU to encode completed state/action
intervals. A feedforward output head combines its history code with the
latest state and calibration inputs $\tilde m,\tilde T$ to reconstruct
the current wind vector. Trained on randomized
collection data, it supplies the unchanged predictor through its wind
input; it never receives the candidate actions it helps evaluate.
This reconstruction recovers the control benefit with no detected
gap to ground-truth wind. Its benefit also persists under the specified noisy,
delayed state interface.

The trajectory-trained GRU learns from closed-loop histories; the reference
GRU uses randomized collection episodes and calibration inputs. The variants
in Table~\ref{tab:wind} thus differ in training data and input interface.
Subsequent comparisons vary only the reference GRU's calibration input
or training.

A separate update test uses the trajectory-trained reconstruction module, seed 0,
in held-out free space at L3. Freezing its reconstructed wind at control
step 20 reduces success from 97 to 59.

\subsection{Shared Calibration in the Reconstruction Path}
\label{sec:recon}
The reference wind reconstruction module receives the same calibration
inputs as the predictor. We first perturb only the reconstruction module's
calibration in held-out obstacles at L3, while the fixed-weight,
wind-conditioned predictor receives correct
calibration. Success decreases under this intervention
(Table~\ref{tab:shared}).

GRU+DCC, the DCC-trained reconstruction variant, preserves the GRU
architecture, loss, and training budget;
calibration perturbations remain fixed per training sample. With the
predictor correctly calibrated, this repair improves success under
reconstruction-only corruption but lowers nominal success. Withholding
calibration yields 55\% in the separate seed-0 control, providing a
reference for that nominal cost.

The reconstruction-only intervention leaves the predictor's direct
calibration input correct. For a shared calibration source, we deliver
the same positive error to both modules, keeping the physical conditions,
predictor weights, and planner fixed. The reference GRU module recovers its nominal mean success,
with no detected net loss relative to correct calibration. GRU+DCC also
remains near its own nominal mean, while performing below the baseline GRU
under shared error. The paired success difference, GRU+DCC minus GRU,
changes sign between the two input conditions (Table~\ref{tab:shared});
all three reconstruction-module seeds show the same reversal.
Across all paired scenes, the GRU+DCC--GRU effect on terminal goal
distance increases by $+0.31$\,m (95\% CI $[0.11,0.54]$) from
reconstruction-only to shared corruption (Table~\ref{tab:shared}).
Success counts completed scenes within the goal tolerance, whereas mean
terminal distance also averages over crashes and timeouts.

\subsection{Reconstruction by a Disturbance Observer}
The DOB replaces the GRU with the same available state, executed actions,
calibration, predictor, and MPPI. It subtracts nominal thrust, gravity,
and drag from acceleration estimated by velocity differencing over each
completed control interval. Inverting the linear aerodynamic map converts
the residual to wind, low-pass filtered with $\alpha=0.5$ fixed before
held-out evaluation. The estimate starts at zero each episode and is
held over the prediction horizon.

The DOB reproduces the baseline GRU module's qualitative response
at $e=+0.1$: success falls under isolated
corruption and recovers when the predictor shares the same error
(Table~\ref{tab:shared}). GRU+DCC's mean success is higher than the
DOB's under reconstruction-only corruption and lower under shared
corruption in the prespecified supplementary comparisons.

\subsection{Calibration-Dependent Compensation}
\label{sec:compensation}
A reconstructed wind vector can absorb error in the model used to
interpret motion. Let $a_0(\eta)$ denote nominal acceleration from
thrust, gravity, and linear drag, and $B_\eta$ the wind-to-acceleration
map at calibration $\eta=(\tilde m,\tilde T)$. For an instantaneous, unfiltered DOB residual
$r_\eta=a_{\mathrm{obs}}-a_0(\eta)$, inversion gives
$B_\eta\hat w_{\mathrm{raw}}=r_\eta$ when $B_\eta$ is invertible. Reusing
the same nominal model and map gives
\begin{equation}
a_0(\eta)+B_\eta\hat w_{\mathrm{raw}}
=a_0(\eta)+r_\eta=a_{\mathrm{obs}}.
\label{eq:compensation}
\end{equation}
This identity isolates compensation before filtering, temporal evolution,
and learned residuals enter the deployed system.

We test this mechanism on fixed recorded histories, comparing the
analytic acceleration $a_0+B\hat w$ with its true-calibration,
true-wind counterpart. Table~\ref{tab:compensation}(a) shows that shared
positive corruption restores GRU and DOB errors to near nominal levels.
GRU+DCC changes less under reconstruction-only corruption and shows much
less recovery under the shared fault. For GRU, the wind-term error alone,
$\|B_\eta\hat w-B_\theta w\|$ with true calibration $\theta$, falls from
1.93 to 1.83\,m/s$^2$ when corruption is shared. Under reconstruction-only
corruption, $\eta=\theta$ in the predictor, so the nominal-dynamics terms
cancel and this equals the combined acceleration error. Under shared
corruption, combining wind with nominal dynamics instead gives
1.00\,m/s$^2$, near the nominal reference of 1.10\,m/s$^2$.

In the predictor, drag is represented through learned residuals with
calibration-dependent inputs. Full 20-step prediction with GRU reconstruction has no detected
error reduction from isolated to shared corruption; Table~\ref{tab:shared} supplies the
closed-loop evidence.

\input{table_compensation}

The negative-offset check retains the same point-estimate ordering but
does not detect either method preference (Table~\ref{tab:compensation}(b)).
Shared-error GRU success stays below its nominal reference: recovery
depends on the reconstruction module's response to the calibration error.

With the downstream predictor fixed, nine combinations of reconstruction
method and calibration interface give descriptive Spearman correlations
with success of $-0.84$ for analytic acceleration error and $-0.72$ for
20-step position error (Tables~\ref{tab:shared} and~\ref{tab:compensation}(a)).
Both errors covary with success here. Table~\ref{tab:main} instead varies
the predictor under correct inputs, where no success-rate difference is detected.

\section{Discussion}
\label{sec:bounds}
\subsection{History Dependence and Online Updating}
\label{sec:mech}
The history probes test whether predictions depend on the history
encoding; closed-loop freezing tests whether updating that encoding
improves control. In separate audits with model seed 0, holding the
history encoding after the first replan yields no detected net success loss for
WM-A+DCC in held-out obstacles at $e=+0.1$ (67 versus 67), or for the
wind-withheld model in selection free space at L3 (86 versus 85).
Current-state feedback and candidate-specific latent updates remain
active.

\subsection{Planning and Perception}
\label{sec:ceiling}
In the no-wind, correct-calibration comparison (model seed 0),
eight configurations spanning the nominal predictors, both ground-truth-clearance controls,
and the two nominal +DCC variants, 63 scenes succeed in every configuration
and 23 fail in every configuration. Ground-truth clearance resolves many collisions but leaves
timeouts. On the selection library, tested changes to
candidate count and use of a mode-preserving sampler did not improve
success.

\subsection{Repair Selection under Shared Inputs}
Table~\ref{tab:record} makes the benefit--cost asymmetry explicit.
For the isolated faults, GRU+DCC needs a much higher fault frequency
than the predictor repairs to offset its nominal cost.
For the shared-fault scenario, its mean success is lower at both
endpoints, so no mixture favors it. The acceleration diagnostic shows
how suppressing a dependency can weaken useful compensation.
Selecting a repair from isolated-fault performance would therefore
misjudge this shared deployment interface.

\subsection{Limitations}
\label{sec:limitations}
Equation~\eqref{eq:compensation} is an instantaneous residual identity
at a common state and action when the wind map is invertible and the same
nominal model is used for inversion and recomposition. It does not identify
physical wind uniquely: unmodeled forces, derivative error, and actuator
mismatch can enter $\hat w$ as an effective disturbance. After low-pass
filtering and holding the estimate over a horizon, the equality need not persist.

The acceleration diagnostic uses a linear analytic map on fixed histories.
The learned predictor represents drag through learned residuals and
calibration-dependent features; its full 20-step rollout does not show a
detected error reduction from isolated to shared corruption. Near-nominal
instantaneous acceleration therefore identifies a compensation route rather
than exact rollout cancellation.

Closed-loop evidence uses one simulated quadrotor and MPPI stack, with
exact or synthetically perturbed state feedback. The DOB uses the same
nominal linear aerodynamic form as the simulated wind channel, a favorable
structural match. Different aerodynamic regimes, state estimators, planners,
and hardware will test when the shared-error compensation and repair reversal
persist. The audit protocol does not require this match, but the mechanism and
numerical repair preference may change without it.

Scene-bootstrap intervals condition on the trained checkpoints and recorded
executions. The DCC noise law and coupled $\pm10\%$ faults are controlled
stress tests rather than measured field distributions. The break-even fault
frequency $p^*$ is a two-condition scenario threshold, not an estimated
deployment failure rate; time-varying and correlated interface errors remain
important transfer tests.

\section{Conclusion}
Correct-input prediction errors span a 12.6$\times$ range with no detected
success-rate difference. Yet fixed-weight input interventions change success
substantially: coupled 10\% calibration errors reduce WM-A from 69\% to 8\%,
and uncertainty training restores 65\%. Supplying zero wind removes the
wind-conditioned model's advantage. Reconstruction recovers control benefit
with no detected success gap to ground-truth wind, while inheriting
calibration dependence.
\par

Under the positive offset, sharing calibration reverses reconstruction
preference. Fixed-history diagnostics reveal that reconstruction bias can
compensate for nominal-model error. Reducing calibration dependence can
weaken this compensation, leaving a repair chosen in isolation inferior
in the connected system.

Repairs should be evaluated under the shared information paths in
which they will operate. A natural extension is to examine how shared
state-estimation errors alter repair selection across reconstruction,
internal prediction~\cite{wolpert1995internal}, and control.

World models should deliver closed-loop benefit from evidence the
robot can own.

\clearpage
\appendices
\section{Implementation and Reproducibility Details}
\subsection{Wind Reconstruction and Conditioning}
The reconstruction module encodes 12 completed 20-Hz intervals, each with
the 18-dimensional state and previous four-dimensional action, using a
single-layer GRU with 32 hidden units. Its output, the latest state, and the
supplied mass and thrust scale enter a 64-unit GELU layer and a linear
three-dimensional wind head. Inputs are standardized from the training pool.
The module trains on 53,344 paired windows with an episode-level 95/5 split,
wind-vector mean-squared error, Adam at $10^{-3}$, batch size 256, and 40
epochs. Seeds 0--2 change initialization and minibatch order. At each replan,
the current estimate is recomputed from completed history and held over the
20-step rollout; future wind is never supplied.

For DCC, the two calibration components receive independent multiplicative
noise from $[-0.2,0.2]$, held fixed within each training window. Physical
state--action samples and targets remain unchanged, and a separate random
stream preserves the reference initialization and minibatch order. The same
rule is applied at the world-model and reconstruction interfaces. The DOB
instead differences velocity, subtracts nominal thrust, gravity, and linear
drag, and inverts the nominal wind map. Its estimate starts at zero and uses
a fixed low-pass coefficient $\alpha=0.5$.

\subsection{Statistical Aggregation}
Comparisons pair scene and disturbance seeds. Single-seed binary outcomes
use exact McNemar tests. For three-seed comparisons, differences are averaged
over trained seeds within each scene before sign tests and scene-bootstrap
intervals. Terminal distance and clearance use all paired scenes; completion
time conditions on joint success. Intervals therefore describe scene
variation for the fixed checkpoints and recorded executions.

\bibliographystyle{IEEEtran}
\bibliography{IEEEabrv,refs}   % 字符串定义文件 IEEEabrv 必须排在 refs 前
\end{document}

%% file: table_main.tex
% Table I: rows follow the common-trajectory accuracy ladder; columns separate
% prediction error from closed-loop outcomes. Requires booktabs only.
\begin{table}[t]
\caption{Prediction-accuracy comparison on the held-out obstacle library (100 scenes,
$v_{\max}$ 3.4\,m/s, learned models use seed 0; success across three seeds is
69--70 for WM-U and 68--71 for WM-A). Error: median 20-step position error on common
trajectories collected with the true-parameter model and on each model's own trajectories.
Time: median over successful runs.}
\label{tab:main}
\centering\footnotesize
\setlength{\tabcolsep}{3.5pt}
\renewcommand{\arraystretch}{1.12}
\begin{tabular}{@{}l rr rrr r@{}}
\toprule
 & \multicolumn{2}{c}{Error (cm)} & \multicolumn{3}{c}{Outcome /100} & \\
\cmidrule(lr){2-3}\cmidrule(lr){4-6}
Model & common & own & Success & Crash & Timeout & Time (s) \\
\midrule
DKI        & 32.8 & 31.0 & 68 & 13 & 19 & 4.25 \\
WM-A       &  8.2 &  9.5 & 69 & 21 & 10 & 2.75 \\
WM-U       &  6.3 &  7.4 & 69 & 22 &  9 & 3.00 \\
True-parameter &  2.6 &  2.6 & 69 & 16 & 15 & 3.20 \\
\addlinespace
\multicolumn{7}{@{}l}{\textit{With ground-truth clearance}} \\
WM-A       &  8.2 &  7.3 & 73 &  4 & 23 & 2.50 \\
True-parameter &  2.6 &  0.6 & 74 &  4 & 22 & 2.95 \\
\bottomrule
\end{tabular}
\par\smallskip\raggedright
Paired time minus WM-A on common successes: DKI $+1.40$\,s
($n=65$); true-parameter $+0.35$\,s ($n=66$), medians of within-scene differences.
\end{table}

%% file: table_shared_calibration.tex
% Reconstruction comparison, PAPER_ICRA sections 13.83 and 13.87.
\begin{table}[t]
\caption{Wind reconstruction on held-out L3 obstacles (100 scenes).
GRU and GRU+DCC average three reconstruction seeds; predictor seed 0 and
DOB are fixed. $(e_{\mathrm R},e_{\mathrm P})$: reconstruction and predictor
calibration errors. CIs bootstrap scene-wise mean differences;
success differences are percentage points (pp). DCC changes reconstruction training.}
\label{tab:shared}
\centering\footnotesize
\setlength{\tabcolsep}{3pt}
\begin{tabular}{@{}lccc@{}}
\toprule
Interface & Correct & Reconstruction-only & Shared \\
$(e_{\mathrm R},e_{\mathrm P})$ & $(0,0)$ & $(+0.1,0)$ & $(+0.1,+0.1)$ \\
\midrule
GRU & 67.0 & 49.3 & 67.0 \\
GRU+DCC & 57.3 & 56.0 & 56.0 \\
DOB & 66 & 44 & 63 \\
\addlinespace
\multicolumn{4}{l}{\emph{GRU+DCC minus GRU (pp):}}\\
Difference & $-9.7$ & $+6.7$ & $-11.0$ \\
95\% paired CI & $[-16.0,-4.0]$ & $[3.0,11.0]$ & $[-17.3,-5.3]$ \\
Wins:losses & $2{:}15/2{:}11/2{:}9$ & $8{:}1/6{:}0/7{:}0$ & $1{:}10/1{:}15/2{:}12$ \\
\addlinespace
\multicolumn{4}{l}{\emph{DOB minus GRU (pp):}}\\
Difference & $-1.0$ & $-5.3$ & $-4.0$ \\
95\% paired CI & $[-4.7,2.3]$ & $[-11.7,0.7]$ & $[-8.7,0.3]$ \\
Wins:losses & $1{:}3/2{:}4/2{:}1$ & $4{:}9/3{:}9/4{:}9$ & $2{:}5/1{:}6/2{:}6$ \\
\addlinespace
\multicolumn{4}{l}{\emph{DOB minus GRU+DCC (pp):}}\\
Difference & $+8.7$ & $-12.0$ & $+7.0$ \\
95\% paired CI & $[3.0,15.0]$ & $[-19.3,-5.0]$ & $[0.7,13.7]$ \\
Wins:losses & $12{:}1/9{:}2/9{:}1$ & $2{:}14/2{:}14/1{:}13$ & $8{:}2/12{:}3/11{:}5$ \\
\midrule
\multicolumn{4}{l}{DOB--GRU interaction\textsuperscript{a}: $+1.3\,[-6.0,8.7]$ pp}\\
\midrule
\multicolumn{4}{l}{\emph{Terminal goal distance: GRU+DCC minus GRU (m):}}\\
Mean difference & $+0.15$ & $-0.05$ & $+0.26$ \\
95\% paired CI & $[0.03,0.26]$ & $[-0.14,0.04]$ & $[0.08,0.46]$ \\
\multicolumn{4}{l}{Shared minus reconstruction-only: $+0.31\,[0.11,0.54]$ m}\\
\bottomrule
\end{tabular}
\par\smallskip\raggedright
\textsuperscript{a}Change in DOB--GRU difference from reconstruction-only to
shared error. DOB--GRU under shared error is the prespecified primary
DOB comparison; supplementary intervals are pointwise.
Terminal distance uses all paired scenes at trial termination; lower is better.
Remaining paired contrasts are in the reproduction record.
Wins:losses are first-method-only:second-method-only successes, in seed
order 0/1/2; DOB is compared with each learned seed.
\end{table}

%% file: table_audit_record.tex
% Joint bootstrap thresholds accompany the canonical success contrasts.
\begin{table}[t]
\caption{DCC decision record on held-out obstacles at $e=+0.1$
(predictors: L0; GRU: L3). Success differences are +DCC minus baseline
under correct ($\Delta_0$) and faulty ($\Delta_f$) inputs;
their CIs appear in Tables~\ref{tab:calib} and~\ref{tab:shared}.}
\label{tab:record}
\centering\footnotesize
\setlength{\tabcolsep}{4pt}
\renewcommand{\arraystretch}{1.12}
\begin{tabular}{@{}lrrr@{}}
\toprule
DCC repair / fault path & $\Delta_0$ (pp) & $\Delta_f$ (pp) & $p^\star$ (\%) [95\% CI] \\
\midrule
WM-A & $-1.0$ & $+57.3$ & $2\,[0,6]$ \\
WM-U & $-1.3$ & $+40.0$ & $3\,[0,8]$ \\
GRU: reconstruction only & $-9.7$ & $+6.7$ & $59\,[33,80]$ \\
GRU: shared with predictor & $-9.7$ & $-11.0$ & No crossing \\
\bottomrule
\end{tabular}
\par\smallskip\raggedright
In a two-condition episode mixture with fault probability $p$, the
expected success difference is $(1-p)\Delta_0+p\Delta_f$.
For $\Delta_0<0<\Delta_f$, the break-even is
$p^\star=-\Delta_0/(\Delta_f-\Delta_0)$; DCC is favored above it.
Unrounded effects are used. Intervals jointly
bootstrap both effects on the same paired scene resamples. Nonnegative
effects at both endpoints give $p^\star=0$. ``No crossing'' denotes lower
DCC success throughout $p\in[0,1]$ at the observed means.
\end{table}

%% file: table_compensation.tex
\begin{table}[t]
\caption{Calibration compensation on 100 held-out L3 obstacle scenes.
(a) Mean of per-seed window medians over three reconstruction seeds;
DOB fixed. The 1,461 common windows were recorded with nominal GRU seed 0.
(b) Success/100, reconstruction and predictor seed 0.}
\label{tab:compensation}
\centering\footnotesize
\setlength{\tabcolsep}{4pt}
\begin{tabular}{@{}lcccc@{}}
\toprule
\multicolumn{5}{@{}l}{\emph{(a) Analytic acceleration error (m/s$^2$)}}\\
$(e_{\mathrm R},e_{\mathrm P})$ & $(0,0)$ & $(+0.1,0)$ & $(0,+0.1)$ & $(+0.1,+0.1)$ \\
\midrule
GRU & 1.10 & 1.93 & 1.84 & 1.00 \\
GRU+DCC & 1.49 & 1.64 & 1.89 & 1.57 \\
DOB & 0.54 & 2.32 & 2.08 & 0.53 \\
\bottomrule
\end{tabular}
\par\smallskip
\begin{tabular}{@{}lccc@{}}
\toprule
\multicolumn{4}{@{}l}{\emph{(b) Closed-loop negative-offset check}}\\
$(e_{\mathrm R},e_{\mathrm P})$ & $(0,0)$ & $(-0.1,0)$ & $(-0.1,-0.1)$ \\
\midrule
GRU & 68 & 56 & 60 \\
GRU+DCC & 55 & 59 & 52 \\
\midrule
GRU+DCC-only:GRU-only & $2{:}15$ & $11{:}8$ & $3{:}11$ \\
Exact McNemar $p$ & $0.002$ & $0.648$ & $0.057$ \\
\bottomrule
\end{tabular}
\end{table}